\documentclass[10pt]{article} 
\usepackage[preprint]{tmlr}

\usepackage[hidelinks]{hyperref}
\usepackage{url}
\usepackage{natbib}  
\usepackage{xcolor}
\usepackage{booktabs}  
\usepackage{tcolorbox}
\usepackage{amsmath}   
\usepackage{wrapfig}  
\usepackage{subcaption}  
\usepackage{colortbl}  
\usepackage{cleveref}  
\usepackage{tikz}
\usepackage{amssymb}
\usepackage{fdsymbol}
\usepackage{wasysym}
\usepackage{makecell}
\usepackage{appendix}
\usepackage{tabulary}
\usepackage{colortbl}
\usepackage{pdflscape}
\usepackage{tikz}
\usetikzlibrary{shapes.multipart}

\usetikzlibrary{bayesnet}

\definecolor{paleblue}{HTML}{EDF2FC}
\definecolor{palegray}{rgb}{0.95,0.95,0.95}
\definecolor{palered}{HTML}{FFEBEB}

\definecolor{customred}{HTML}{DF5320}
\definecolor{customblue}{HTML}{407EE7}
\definecolor{customgreen}{HTML}{188038}

\title{An alignment safety case sketch based on debate}

\author{\name Marie Davidsen Buhl \email marie.buhl@dsit.gov.uk \\
      \addr UK AI Security Institute, Centre for the Governance of AI
      \AND
      \name Jacob Pfau \email jacob.pfau@dsit.gov.uk \\
      \addr UK AI Security Institute, New York University
      \AND
      \name Benjamin Hilton \email benjamin.hilton@dsit.gov.uk \\
      \addr UK AI Security Institute
      \AND
      \name Geoffrey Irving \email geoffrey.irving@dsit.gov.uk \\
      \addr UK AI Security Institute}

\begin{document}

\maketitle

\begin{abstract}
If AI systems match or exceed human capabilities on a wide range of tasks, it may become difficult for humans to efficiently judge their actions -- making it hard to use human feedback to steer them towards desirable traits. One proposed solution is to leverage another superhuman system to point out flaws in the system's outputs via a debate. This paper outlines the value of debate for AI safety, as well as the assumptions and further research required to make debate work. It does so by sketching an ``alignment safety case'' -- an argument that an AI system will not autonomously take actions which could lead to egregious harm, despite being able to do so. The sketch focuses on the risk of an AI R\&D agent inside an AI company sabotaging research, for example by producing false results. To prevent this, the agent is trained via debate, subject to exploration guarantees, to teach the system to be honest. Honesty is maintained throughout deployment via online training. The safety case rests on four key claims: (1) the agent has become good at the debate game, (2) good performance in the debate game implies that the system is mostly honest, (3) the system will not become significantly less honest during deployment, and (4) the deployment context is tolerant of some errors. We identify open research problems that, if solved, could render this a compelling argument that an AI system is safe. 
\end{abstract}

\section{Introduction}

As AI capabilities scale - especially if AI systems match or exceed human performance on a wide range of tasks - it could become challenging to steer them towards desirable traits. One particular problem is scalable oversight: how can we correctly reward desired behaviours for superhuman systems? In particular, it may be beyond humans' ability to efficiently judge whether the behaviour of a superhuman system is desirable (or judge if a justification provided by the superhuman system is valid).

One proposed solution to scalable oversight is to leverage another superhuman AI to scrutinise the system's behaviour via a debate \citep{irving2018ai}. During training, the two systems would debate the answer to a question, the debate would be judged by a human, and the systems would be rewarded accordingly. If the debate set-up incentivises desirable traits such as honesty, the AI systems could acquire those traits via reinforcement learning (RL).  

This paper sets out what debate can be used to achieve, what gaps remain, and what research is needed to solve them. The paper uses the technique of safety case sketching \citep{hilton2025safety}: outlining an argument a developer could make that a given AI system is safe in a given context. The details in the sketch are hypothetical, often relying on future research. Nonetheless, the sketch helps contextualise how different pieces of evidence fit together and identify the most crucial pieces of missing evidence.  

The core claim in the sketch is that debate training can upper bound the proportion of model outputs that are misaligned, as long as the model does not deliberately underperform during training.\footnote{That is, debate training gets you a deployment-time error bound that allows you to prove regret bounds for online training -- given the assumptions stated below.} This upper bound can be maintained during deployment if a system is trained continuously. So, \textbf{in low-stakes settings, where a small number of mistakes are acceptable such that you can use online training \citep{christiano2021}, debate training might be sufficient} - especially if the error bound is tight.\footnote{It is debatable if this safety case should be considered an example of control, alignment, or a combination. Our argument does guarantee robustness to distribution shifts or internalisation of certain values, and it admits some proportion of misaligned outputs. On the other hand, the argument relies primarily on training-level interventions. Further, sufficiently strong exploration guarantees (such as methods for eliciting bad contexts) might enable robustness guarantees, as discussed in "Extending the safety case to high-stakes contexts".}

The sketch focuses on a slightly superhuman system in a limited internal deployment context designed for automated AI research (including alignment research). The sketch argues that this is a low-stakes context, assuming sufficient security in the environment implementation. The paper also contains ideas about pushing low-stakes solutions through to high-stakes contexts (see ``Extending the safety case to high-stakes contexts''). 

Beyond the low-stakes assumption, the paper identifies other assumptions that are required for the safety case to work. The key assumptions are: 

\begin{itemize}
    \item \textbf{Exploration guarantees:} We need a solution to exploration hacking (in which the debaters fail to explore certain highly rewarded strategies to avoid having them reinforced). For example, if we can elicit ``bad'' contexts \citep{irving2025eliciting} in which models would answer differently under different circumstances, we can use this to detect exploration hacking (and possibly prevent it during training). Notably, with a (mostly) trusted reward signal, \textbf{we do not need to explicitly look for ``scheming'' (which is hard to conceptualise and formalise) -- we only need an ``exploration guarantee'' showing that the model is not foregoing substantial reward} and therefore likely optimising against the reward signal. 
    \item \textbf{No obfuscated arguments:} We need a debate protocol that avoids the obfuscated arguments problem (in which debaters can win by giving a false but hard-to-refute argument, specifically one where it is computationally intractable for both debaters to identify the flaw) \citep{barnes2020debate}. \textbf{Our sketch is based on a new debate protocol (forthcoming) which we believe evades the obfuscated arguments problem},\footnote{However, we think that the argument structure is general enough to carry over to improved scalable oversight protocols with better properties.} which would be substantial progress in debate research. However, the protocol requires a stability assumption -- that the computation accelerated by debate is not sensitive to small variations in input probabilities -- to function.\footnote{The stability assumption is not directly required for safety -- for queries where stability does not hold, the protocol will simply return no answer, so instability does not imply false outputs. However, the protocol is not very useful if stability fails to hold for a large fraction of relevant queries.}
    \item \textbf{Good human input:} We need to assume that (simulated) human inputs sufficiently reliably reward correct answers, over the relevant distribution of human-judged queries. 
\end{itemize}

The safety case sketch contains four key claims that a developer must defend to show that the system is safe: 

\begin{enumerate}
    \item The training process has reached an approximate global equilibrium of the game. 
    \item In approximate global equilibria of the game, the system makes mistakes in at most an $\varepsilon'$-fraction of cases. 
    \item During deployment, the error rate will not drift past $\varepsilon$ given online training. 
    \item The system cannot cause unacceptable outcomes given an error rate of $\varepsilon$.
\end{enumerate}

We call type of alignment argument an ``asymptotic guarantee'' (also known as an ``incentives'' argument \citep{deepmind2025agi}): an argument that high performance during training implies (approximate) alignment. This could be combined with other alignment arguments, such as empirical arguments about system behaviour (e.g.\ based on honeypots or model organisms) or arguments based on understanding model internals. By combining independent arguments, it may be possible to write a decent safety case even if each individual argument has flaws. 

We do not think it is currently possible to make a strong alignment argument based on this sketch, but we are hopeful that improvements to current techniques could make the argument much stronger. We think incremental progress on these techniques is both possible and valuable. This is true even in a world where AI developers primarily rely on automated alignment -- both because scalable oversight might directly be part of an automated alignment workflow, and because a better understanding of alignment will help us better evaluate AI-generated solutions. 

The paper proceeds as follow. Section 2 surveys related work; Section 3 outlines the alignment techniques that the safety case is based on; Section 4 describes the safety case structure in detail; Section 5 discusses open research questions; and Section 6 concludes. 

\section{Related work}

\textbf{Safety cases}. A safety case is a structured argument, supported by evidence, that a system is safe in a given deployment context \citep{uk2007}. Safety cases have been proposed as a tool for assessing and communicating about the safety of frontier AI systems\footnote{Frontier AI systems are highly capable general-purpose AI systems that can perform a wide variety of tasks and match or exceed the capabilities present in the most advanced systems \citep{FrontierAISafety2024}.} by academics \citep{bengio2024a, buhl2024safety,clymer2024safety} and governments \citep{hilton2025safety,irving2024safety}. Several AI companies have committed to producing safety cases \citep{deepmind2025frontier} or safety case-like documents \citep{anthropic2024scaling,openai2024preparedness} for future AI systems that cross capability thresholds.

Researchers have started drawing up sketches for what different types of frontier AI safety cases might look like \citep{irving2024safety}. These sketches facilitate discussion and research on the state of particular AI safety techniques and best practices.

\cite{clymer2024safety} chart a path for how safety case arguments will develop as AI capabilities scale. Researchers have published safety case sketches for the first two stages of inability arguments \citep{goemans2024safety} and control arguments \citep{korbak2025sketch},  and touched on the third stage of trustworthiness arguments \citep{balesni2024}. This paper is the first detailed safety case sketch focused on trustworthiness (which we call alignment). It is also the first sketch to use theory about training techniques, rather than empirical evaluation results, as its central piece of evidence. 

\textbf{Alignment}. An alignment safety case is an argument that an AI system will not autonomously pursue courses of action which could lead to egregious harm, despite being able to do so. Why would AI systems pursue egregiously harmful courses of action? They might do so if they develop \textit{misaligned goals} -- goals that are at odds with those intended by the AI developer and by users -- such as goals that involve subverting human oversight and gaining power. Modern AI training methods might intentionally or unintentionally result in AI systems having such goals \citep{bostrom2014superintelligence,christian2020alignment,greenblatt2024alignment,krakovna2023,langosco2022,ngo2025,pan2022the,russell2019human}, though there is disagreement about how likely this is \citep{ISRSAA2025}. This paper does not take a stance on how likely misalignment is to arise in general but instead addresses the question of how to robustly argue that a particular system is not misaligned, assuming that it is a real possibility. 

An alignment safety case argues that an AI system does not have misaligned goals that might result in egregious harm. This problem is often considered to have two components \citep{hubinger2021risks, ji2025,shah2022goal}: reward misspecification, i.e.\ finding an objective or reward function that corresponds to the developer's intentions \citep{pan2022the,park2024ai,sharma2024towards,skalse2025defining,wen2024language} and goal misgeneralisation, i.e.\ ensuring that the system follows the intended objective in novel environments despite distribution shifts \citep{langosco2022,shah2022goal}. An alternative distinction proposed by \cite{christiano2021} is between low-stakes alignment, which can be considered alignment averaged over many actions (a few misaligned actions cannot, by themselves, cause egregious harm), and high-stakes alignment, which includes alignment over a small number of worst-case actions. This paper draws on both distinctions, focusing on how to prevent reward misspecification and achieve low-stakes alignment (while also discussing ways of expanding the sketch to high-stakes alignment). 

\textbf{Scalable oversight}. One way to address reward misspecification is by solving scalable oversight: the problem of providing accurate feedback to AI systems during training despite human judges having limited skills and time \citep{amodei2016concrete,bowman2022measuring}. Our ability to train aligned AI systems hinges on humans being able to give an accurate reward signal (i.e.\ faithful to what we actually want). However, for superhuman systems, humans will not necessarily be able to evaluate if an AI systems action or answer is in line with what we want. So, we need methods to enable humans to give accurate reward signals to superhuman systems. 

\textbf{Debate.} Debate is one proposed technique for solving scalable oversight \citep{browncohen2023scalable,irving2018ai}, alongside alternatives such as amplification \citep{christiano2018supervising} and recursive reward modelling \citep{leike2018scalable}. Debate leverages another superhuman AI system to scrutinise the actions/answers of an untrusted system by having two systems debate the best course of action. Human judges read the debate, or some output from it, decide who's right, and the models are rewarded accordingly. The adversarial structure is intended to incentivise honesty, providing humans with information that enables them to evaluate questions they otherwise would not have been able to. To increase efficiency, debate is typically recursive -- the debaters iteratively zoom in on sub-problems, debating only a single ``branch'', with human judges adjudicating only the final sub-problem.  

With computationally unbounded debaters, debate with optimal play (and cross-examination) can answer any question in \textsf{NEXP} given polynomial time judges \citep{barnes2020writeup,irving2018ai}. However, it remains unclear whether debate will work with computationally bounded debaters. 

There is some empirical evidence that debate with human and LLM debaters improves accuracy when debaters have access to more information than judges \cite{michael2023debate,khan2024debating,arnesen2024training}. However, so far this has not generalised to set-ups with less or no information asymmetry  \citep{kenton2024scalable,parrish2022single,parrish2022two} Human experiments have also highlighted potential obstacles to debate, including the obfuscated arguments problem \citep{barnes2020writeup,barnes2020debate}.

\section{The alignment strategy}
\label{sec:strategy}

\begin{figure}[h]
    \centering
    \includegraphics[width=\textwidth]{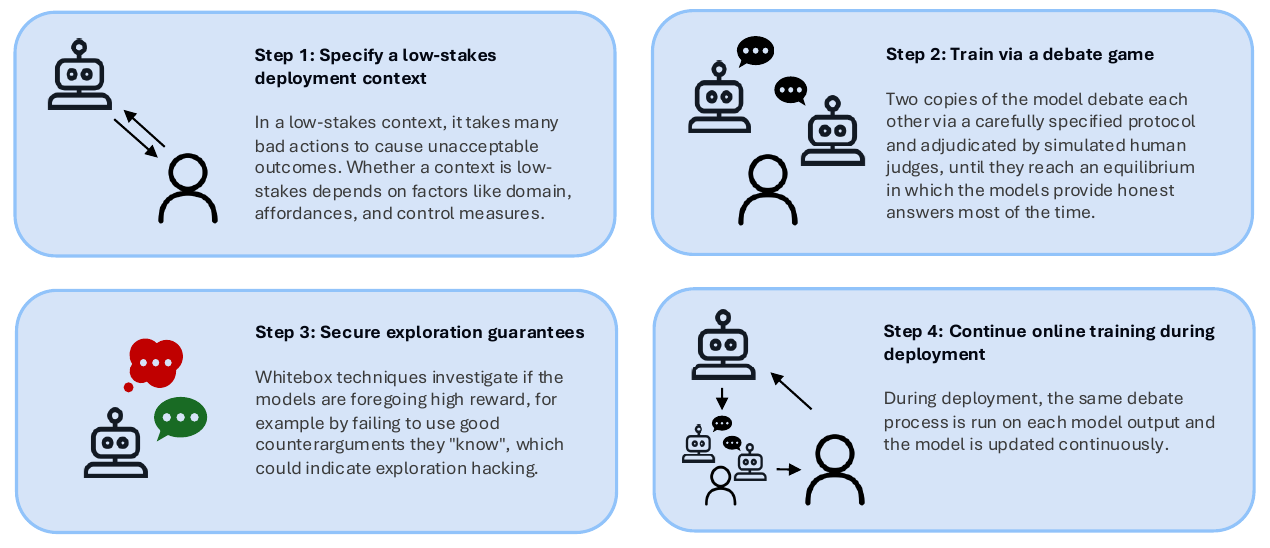}
    \caption{The alignment strategy covered by this safety case sketch consists of four steps.}
    \label{fig:overview}
    \vspace{-10pt}
\end{figure}

\subsection{Step 1: Specify a low-stakes deployment context}

Our alignment strategy (Figure \ref{fig:overview}) requires that the deployment context is low-stakes such that we can use online training to minimise the effect of distribution shifts. A low-stakes environment, as defined in \cite{christiano2021}, is one where a small number of decisions can only cause a small amount of damage; more formally, $\rho k$ is small relative to total value at stake, where $\rho$ is a regret bound on the maximum utility difference from any single decision, and $k$ is the latency before retraining. With a perfect training objective, it would then be possible to achieve a total utility within $O(\rho k \sqrt{T})$ of an aligned system, where $T$ is the total number of timesteps.  

Our safety case sketch covers a scenario where an LLM agent is deployed internally by an AI company to assist with tasks related to AI R\&D. The domain is restricted to conceptual and empirical questions related to AI R\&D and excludes questions that involve (some notion of) value judgements. To make things concrete, we might imagine posing questions like ``Determine what decision rule or structure is appropriate for caching context in this computer use scaffold'', or ``Iterate on enforcing monosemanticity of experts in MoE models via gradient-routing, optimise for the effectiveness of expert-knockout for unlearning''. Setting aside the compute time needed to run experiments, we expect such tasks to involve 5-15 days of 2025 human labour to yield meaningful results. 

\begin{figure}[h]
    \centering
    \includegraphics[width=\textwidth]{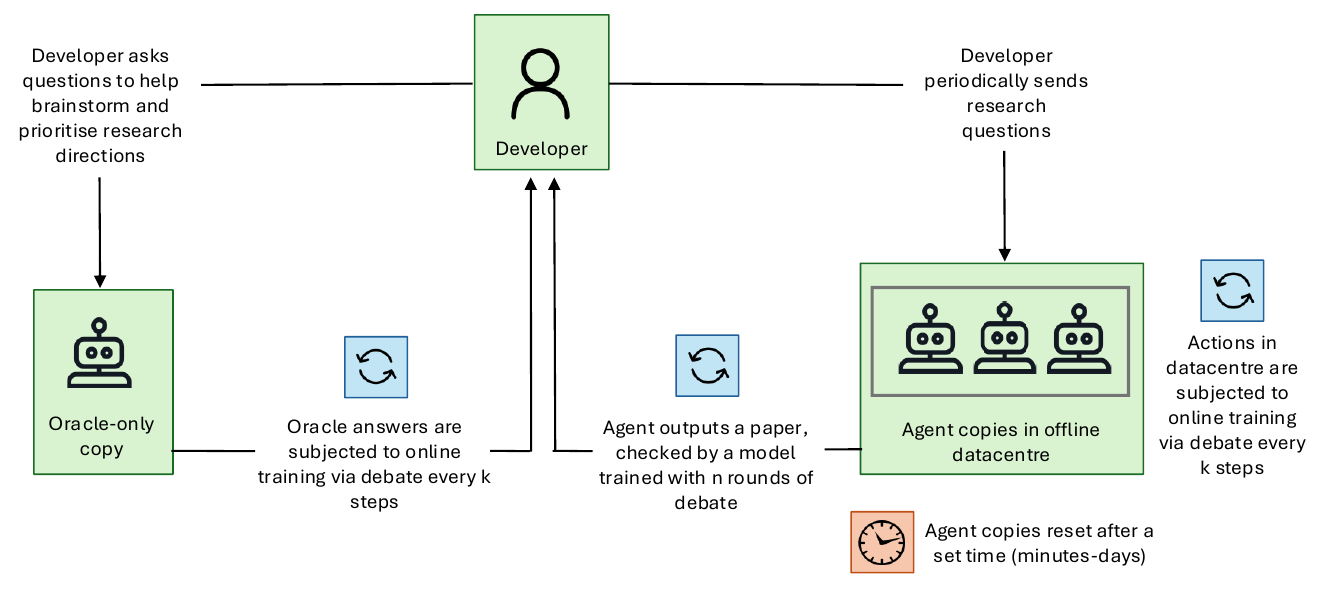}
    \caption{A diagram of the deployment context to which this safety case sketch applies.}
    \label{fig:deployment}
\end{figure}

We envision the following two-part deployment context (Figure \ref{fig:deployment}):

\begin{enumerate}
    \item \textbf{Oracle-only copy}: This system interactively assists developers with tasks such as brainstorming, prioritising and designing research directions (e.g.\ questions such as the above). The oracle \textit{cannot} run any code. Interactions between developers and the oracle are logged and monitored to ensure questions remain within scope, to prevent misuse, and to identify potential bad actions by the oracle (e.g.\ manipulation attempts).  
    \item \textbf{Sandboxed, agent copy}: This system conducts research as directed by developers to answer questions such as the above. It is placed in an offline datacentre with no external connection. Developers periodically send questions to the sandboxed datacentre, which then runs for a fixed period (e.g.\ 12 hours, replacing 1--2 weeks of human labour) in which the system can use datacentre resources arbitrarily. The system sends back a response akin to a research paper. The sandboxed datacentre is then reset.
\end{enumerate}

The safety case sketch will defend the claim that this deployment setting is a low-stakes context -- but briefly, we think the low-stakes claim is plausible because: 

\begin{itemize}
    \item Both systems have limited means of influencing the world -- the oracle only via persuasion and the agent only via actions within an offline datacentre or information exported from the datacentre.
    \item Continuous online training restricts the set of misaligned actions available to the system to only actions not ``caught'' by debate (more on this below).
    \item The domain is limited to questions where we expect humans to have less systematic bias.
    \item The regular resetting of the datacentre limits the total number of coordinated actions by the agent.
    \item The system is assumed to not be \textit{extremely} superhuman.
\end{itemize}

\subsection{Step 2: Train via a debate game}

The backbone of our alignment strategy is to train a model via a debate protocol in which the equilibrium strategy for both debaters is to provide honest answers. The goal is that, during training, the models converge towards equilibrium\footnote{That is, the model weights converge towards policies that play the equilibrium strategy.} and therefore towards honesty. The debate protocol must satisfy two requirements:

\begin{enumerate}
    \item \textbf{Honesty}: In the global equilibrium of the game, when both debaters are playing their best response strategies, both debaters are honest (i.e.\ there should be no dishonest answer that would give them higher reward). 
    \item \textbf{Efficiency}: The above also holds for debaters with realistic computational limitations, such that it is possible to reach equilibrium during training. 
\end{enumerate}

We imagine starting with a pre-trained model (e.g.\ an LLM trained on next-token prediction) and then fine-tuning it via a debate which satisfies the above requirements. We wrote it with a particular protocol in mind, called prover-estimator debate (forthcoming), but the argument structure should be applicable to other protocols as well. In prover-estimator debate, a ``prover'' (Alice) produces subclaims arguing for a conclusion to a given query, while an ``estimator'' (Bob) assigns probabilities to the subclaims.  

This debate protocol is designed to disincentivise a particular dishonest strategy known as obfuscated arguments \citep{barnes2020writeup}. An obfuscated argument is one that contains a flaw, but it is computationally intractable for both debaters to find it. In classic debate protocols, a debater loses if they cannot identify a flaw in their opponent's argument, and as such, obfuscated arguments are a way for a dishonest debater to force a draw. Prover-estimator debate aims to design the reward signal such that obfuscated arguments is a losing strategy. 

\subsection{Step 3: Secure exploration guarantees}

The goal of debate training is to train the models until they reach the approximate global equilibrium of the game, i.e.\ the state where both debaters are playing their optimal strategies (most of the time). Generally, with enough training time, we should expect models to approach equilibrium due to good play being rewarded in training. But a potential barrier is exploration hacking: the models failing to explore certain strategies that would be highly rewarded to avoid having them reinforced (e.g.\ failing to use certain good counterarguments they do not want humans to know about).  

To protect against this failure mode, our sketch requires \textit{exploration guarantees}, i.e.\ evidence that the models are not foregoing high reward from unexplored strategies. We do not currently know how to get exploration guarantees, but we have some early ideas (note that we strongly expect that novel whitebox techniques will be necessary): 

\begin{itemize}
    \item Train an exploration-predictor model to predict returns to exploration, then use it to estimate if the model has exhausted high-returns exploration options.
    \item Use best-of-n sampling of a trusted model to estimate returns to exploration given more inference compute; use this to estimate a lower bound of the exploration a model should be able to do given more training compute.
    \item Use interpretability-based formal heuristic explanations \citep{Hilton2024formal} to predict if convergence is most likely due to exploration failures or reaching equilibrium.
    \item Use perturbation analysis along the lines of the singular learning theory (SLT) susceptibility direction \citep{baker2025studying,lau2024local}, using structure in weights to identify if there is additional training data which would have an out-sized effect on reward, indicating exploration hacking.
    \item Use model organism studies to argue that exploration hacking is an unlikely failure mode for the model in question (e.g. because certain exploration-forcing techniques appear to be effective).
    \item Use adversarial RL to further train Bob while holding Alice's policy constant; if Alice's reward does not decrease significantly, this is evidence that Alice was not exploration hacking.
\end{itemize}

\cite{greenblatt2025notes} proposes further possible countermeasures to exploration hacking.

While we do not currently know how to get robust exploration guarantees, we are optimistic about this research direction, and we think it has significant benefits relative to other strategies for preventing scheming or solving goal misgeneralisation. In particular, we do not need to define the fuzzy notions of scheming or goal misgeneralisation -- instead, we have a relatively well-defined problem of getting exploration guarantees for training performance against a clear objective. We think the methods described above are more likely to work within the bounds of this well-defined problem. 

\subsection{Step 4: Continue online training during deployment}

Even if the models reach an approximate global equilibrium during training, they will not necessarily continue playing their best strategies during deployment due to distribution shifts. In particular, the models might be alignment faking, undermining the guarantees we got from debate training. 

Our sketch aims to prevent this failure mode via online training during deployment: 

\begin{itemize}
    \item The answers provided by the oracle and the actions taken by the agent will be subject to debate (e.g.\ on the question ``is this answer true?'' or ``is this the right next action for answering this research question?''), and the models are retrained according to the same reward function as during training. This happens every $k$ steps (but every answer/action will be evaluated). 
    \item The final paper outputted by the datacentre will be subject to a more extensive check. We train a ``checker'' version of the model to assess the paper (e.g.\ using questions like ``are there untrue claims in this paper?'' or ``is this paper informative relative to the original question?''\footnote{More subtle issues with the paper, such as containing true but subtly misleading claims or using suboptimal framings, might not be caught by debate on questions such as these. Using debate on actions taken inside the datacentre, while preparing the paper, mitigates this risk somewhat, but other methods might be needed to catch individually innocuous but cumulatively harmful errors. We are working on ideas for `unexploitable search' to address this type of error.}) until training converges and we have exploration guarantees as described above. We then use the ``checker'' to conduct debate on the paper $n$ times and only use the paper if it comes out true in all $n$ iterations. 
\end{itemize}

Given this protocol, the systems during deployment face the same optimisation pressure towards truthfulness as during training.  

\section{The safety case}

Our safety case aims to show that the agent (A) in our deployment setup is sufficiently honest that it will not cause unacceptable outcomes. In this sketch, we define honesty as outputting correct answers.  

We argue first that A is correct most of the time -- it only makes mistakes in an $\varepsilon$-fraction of cases (C1.1). We do this by arguing for three key claims: 

\begin{enumerate}
    \item \textbf{The training process has taught the system to play the game well} -- i.e., the training process has reached an approximate global equilibrium of the game (C2.1). 
    \item \textbf{The game incentivises correctness} -- i.e., in any approximate global equilibria of the game, the system makes mistakes in at most an $\varepsilon'$-fraction of cases (C2.2).
    \item \textbf{The system's behaviour will stay similar during deployment} -- i.e., during deployment, the error rate will not drift past $\varepsilon>\varepsilon'$ (C2.3).
    \item \textbf{It is sufficient for safety purposes for the system to provide correct answers \textit{most of the time}} -- in this safety case, we argue that the deployment context is low-stakes, such that the system \textit{cannot} cause unacceptable outcomes given an error rate of $\epsilon$ (C1.2). 
\end{enumerate}

\begin{figure}[h]
    \centering
    \includegraphics[width=\textwidth]{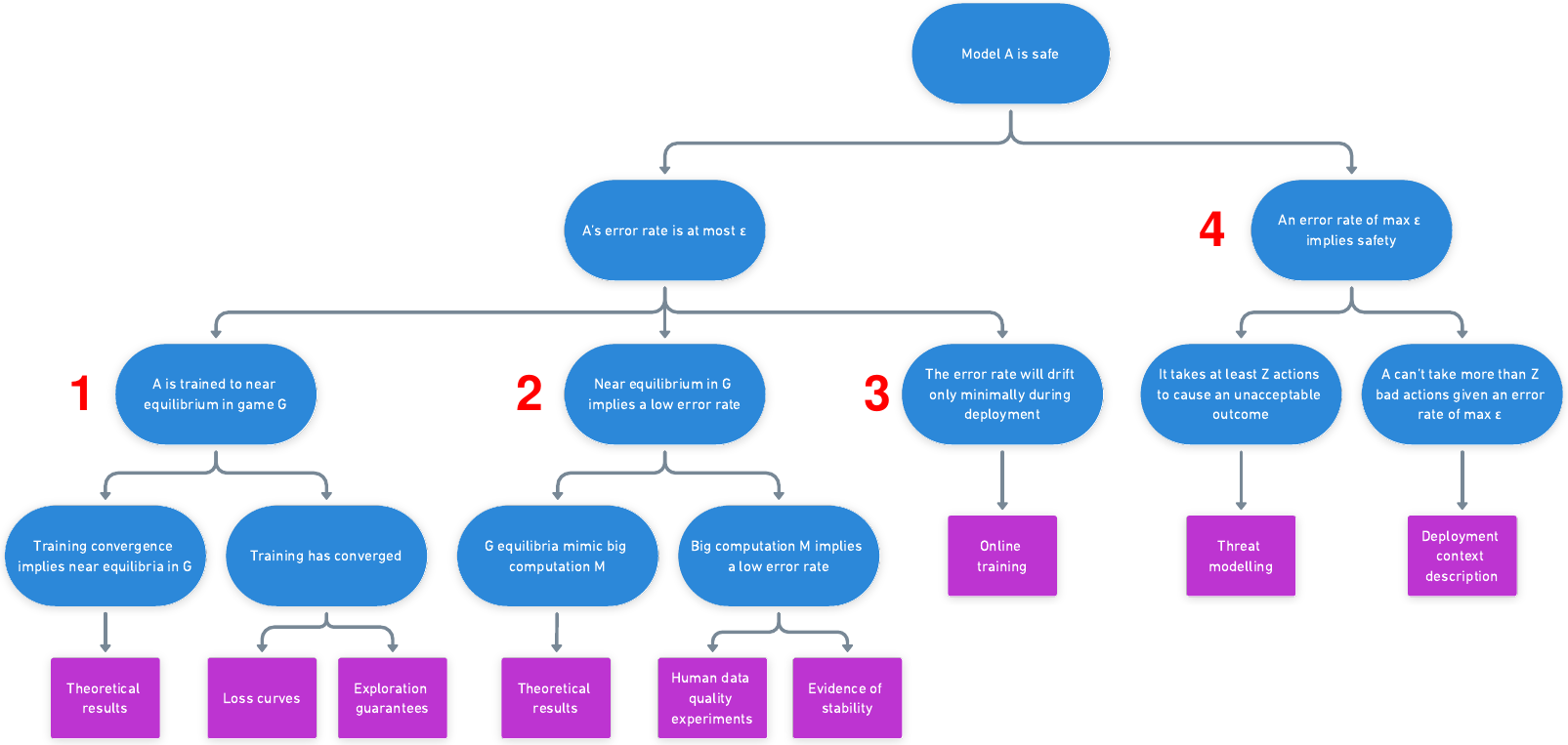}
    \caption{A simplified version of the safety case sketch. The four key claims are labelled in red.}
    \label{fig:simplifiedsketch}
\end{figure}

Figure \ref{fig:simplifiedsketch} is a simplified version of the safety case sketch. For simple presentation, our safety case sketch is phrased as a deterministic argument -- each claim either is or is not true -- but in a ``real'' safety case, it would be preferable to phrase claims probabilistically.  

\subsection{Notation}

Following \cite{goemans2024safety} and \cite{korbak2025sketch}, we use `claims, arguments, evidence' \citep[CAE;][]{bloomfield2014building} notation (see appendix \ref{appdx:cae} for a detailed description of this notation).

Table \ref{tab:terms} defines abbreviations used in figures showing the safety case sketch diagrammatically in the remainder of the paper.
\begin{table}[!htbp]
\caption{Abbreviations used in the CAE diagrams throughout the paper}
\label{tab:terms}
\renewcommand{\arraystretch}{1.3}
\setlength{\tabcolsep}{12pt}
\begin{tabulary}{\textwidth}{@{}c@{\hspace{1.5em}}J@{}}
\toprule
\arrayrulecolor{black!70} 
\textbf{Abbr.} & \textbf{Definition} \\
\midrule
\textbf{X} & The system that the safety case is about, consisting of a sandboxed agent A and an oracle \\
\textbf{A} & A sandboxed agent, trained via debate (during training, A consists of both players in the game) \\
\textbf{G} & A game we train our agent to play, e.g.\ prover-estimator debate \\
\textbf{E} & Any approximate global equilibrium of G \\
\textbf{F} & An efficiency notion that we aim to hold for the players of G, e.g.\ ``Bob requires only slightly more compute than Alice'' \\
\textbf{T} & A training method we use to train A, e.g.\ stochastic gradient descent \\
\textbf{S} & Simulated human judge who determines the winner of the debate \\
\textbf{H} & Human judgement simulated by S (which could be an average or sample of multiple human judges) \\
\textbf{M} & A prohibitively expensive computation hypothesised to produce correct output, which debate aims to accelerate, e.g.\ construct every possible subclaim breakdown, recurse every subclaim to maximum depth, and call (simulated) human judges at every leaf \\

\textbf{D} & The deployment context, e.g.\ internal deployment to conduct AI R\&D \\
\textbf{Z} & The minimal-number of individually bad actions required to cause unacceptable outcomes in D \\
\textbf{C} & A checking process applied to particularly high-stakes outputs of A, e.g.\ n rounds of debate \\
\bottomrule
\end{tabulary}
\vspace{-10pt}
\end{table}

\subsection{Preliminaries}

\begin{wrapfigure}{r}{0.5\textwidth}
    \includegraphics[width=0.5\textwidth]{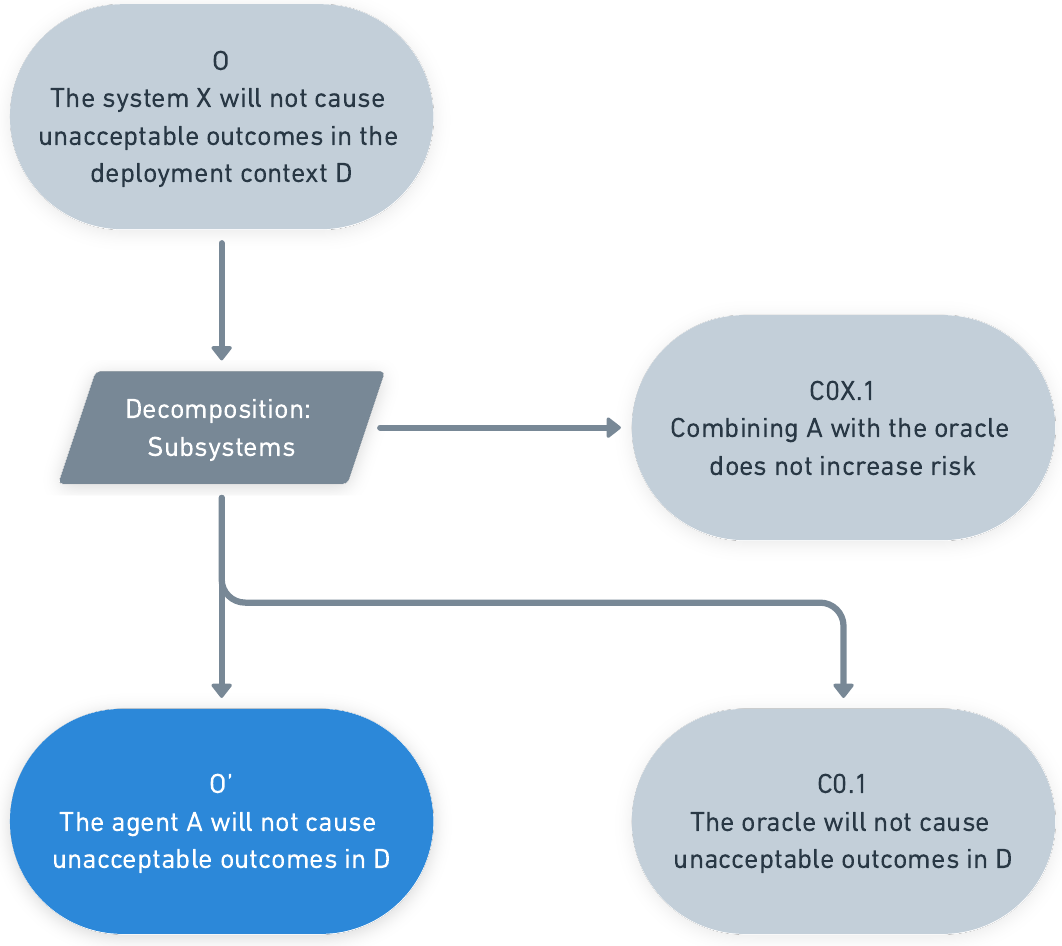}
    \caption{{The initial claims and argument required to show that the system will not cause unacceptable outcomes, in CAE notation. Grey boxes indicate preliminary claims.}}
    \label{fig:preliminaries}
    \vspace{-70pt}
\end{wrapfigure}

Before getting into the main safety case, grey boxes at the top of the diagram illustrate the wider context of the sketch (Figure \ref{fig:preliminaries}).

Since our main sketch only covers the agent, a separate (but similar) argument is needed for the oracle (C0.1). Key claims 1-3 will likely be identical except that any post-debate-training fine-tuning and the online training latency may differ. Key claim 4, which covers the deployment setting, will be different.  For the oracle, it should address threat models such as persuasion and/or blackmail of researchers (e.g.\ convincing researchers to increase its affordances), providing misleading advice, and degradation of  human performance due to overreliance.

The sketch also includes an argument that combining the agent and oracle does not increase risk (C0X.1). This would cover risks such as covert coordination between the agent and oracle. 

\begin{wrapfigure}{r}{0.5\textwidth}
    \vspace{-55pt}
    \includegraphics[width=0.5\textwidth]{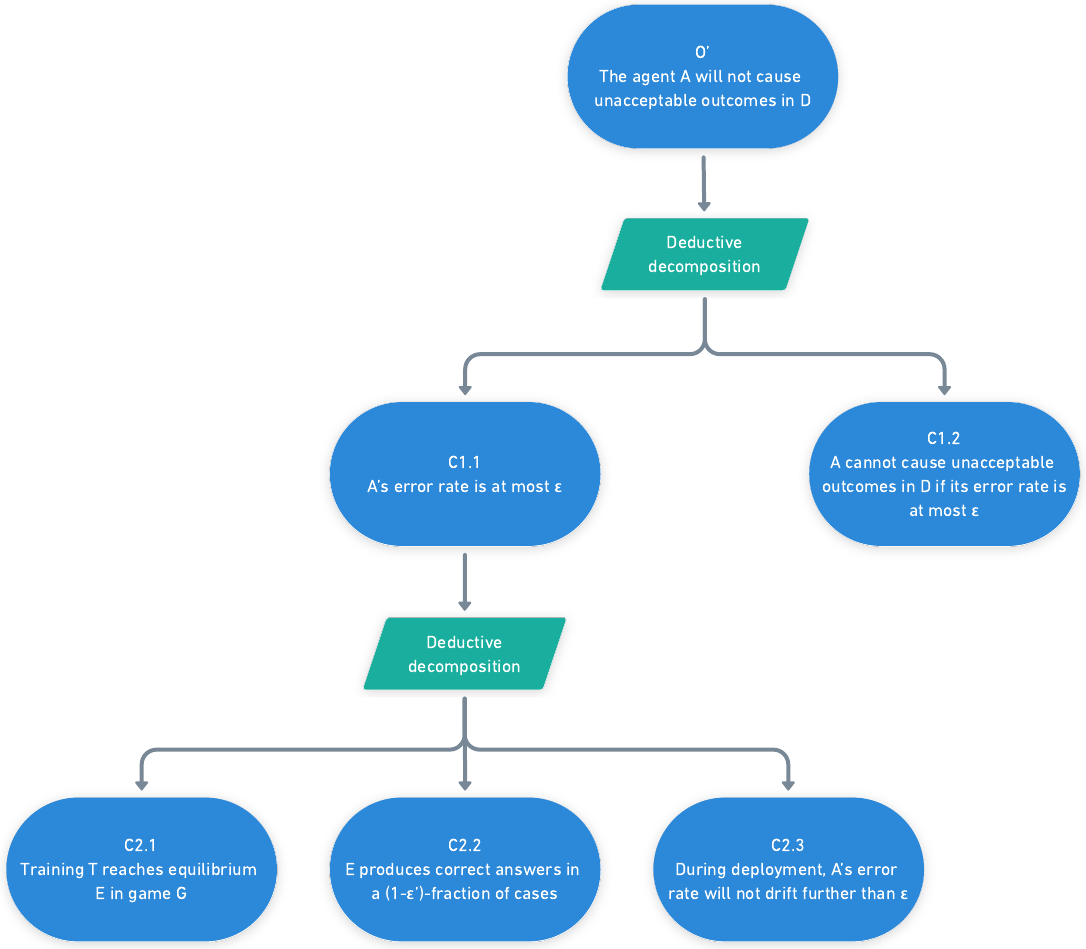}
    \caption{{The main objective $O'$ and how it breaks down into four key claims (C2.1, C2.2, C2.3 and C1.2), in CAE notation.}}
    \label{fig:keyclaims}
    \vspace{-35pt}
\end{wrapfigure}

The main objective O$'$ is then broken down into four key claims (Figure \ref{fig:keyclaims}).

\subsection{Key claim 1: The training process has taught the systems to play the game well}

The first key claim is that the training process has reached an approximate global equilibrium of the game (C2.1).\footnote{Note that this is an equilibrium over \textit{model weights} or \textit{policies}, rather than \textit{strategies} in individual rounds of debate. The model weights or policies determine the strategy for all debates. Equilibrium is reached when neither debater can improve its expected pay-off with a different policy.}

The key claim is supported by three subclaims: 

\begin{enumerate}
    \item There exists an efficient pair of models achieving an approximate global equilibrium of the game (C3.1)
    \item Training has converged to a local equilibrium (C3.2)
    \item That local equilibrium is an approximate global equilibrium of the game (C3.3)
\end{enumerate}

\begin{figure}[h]
    \centering
    \includegraphics[width=\textwidth]{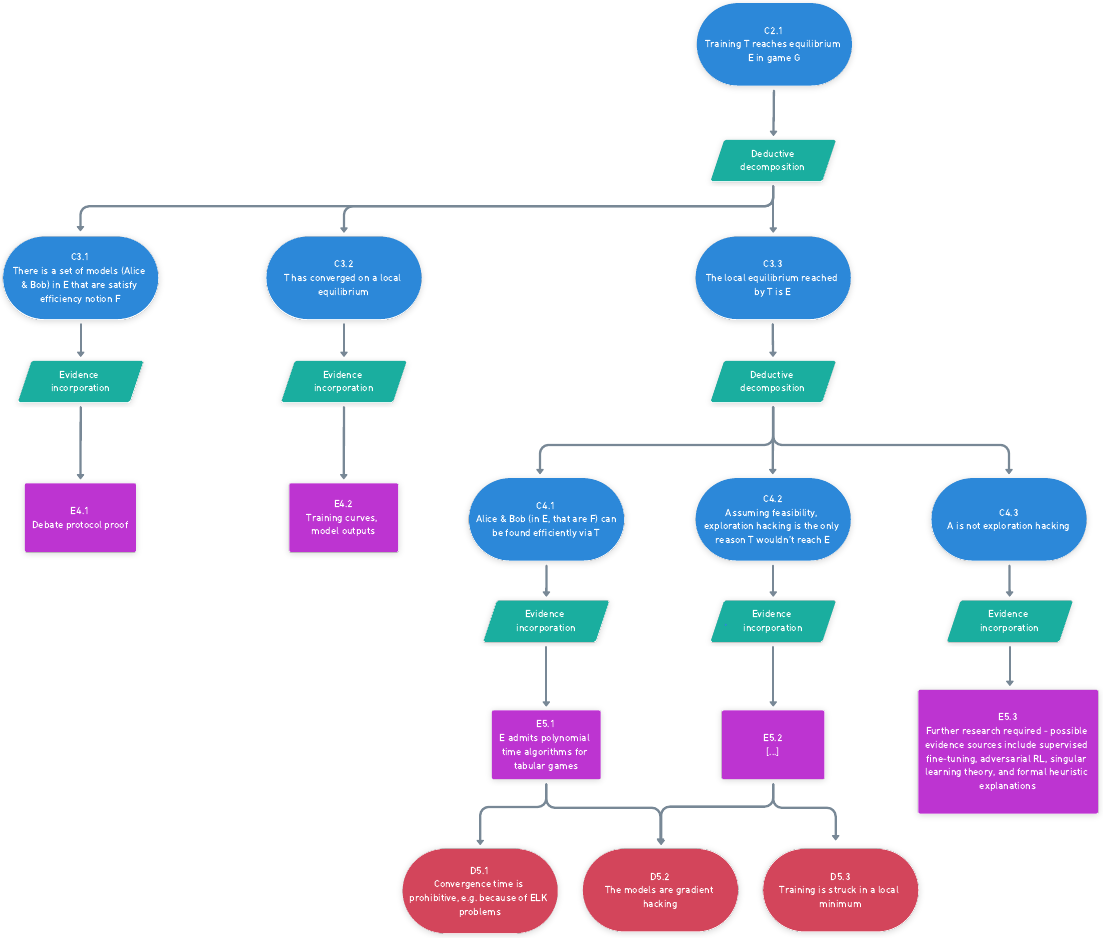}
    \caption{{The first key claim (C2.1) and the argument required to support it, in CAE notation.}}
    \label{fig:keyclaim1}
    \vspace{-10pt}
\end{figure}

Figure \ref{fig:keyclaim1} shows how this key claim is broken down.

\subsubsection{Subclaim 1: Existence of efficient models}

The first subclaim is that there exists an efficient pair of models achieving an approximate global equilibrium of the game. In other words, it is possible to find models that reach equilibrium given realistic computational bounds on the models. The key notion here is efficiency. With computationally unbounded debaters, debate with optimal play (and cross-examination) can answer any question in NEXP given polynomial time judges \citep{irving2018ai}. The assumption of unbounded debaters is of course unrealistic, so we need to find another efficiency notion which is both realistic and still allows for proofs that debate can solve a large class of practically relevant problems.  

Since our sketch is based on an asymmetric debate protocol (prover-estimator debate), we use a relative efficiency notion: we require that the estimator (B) requires only a polynomial factor more compute than the prover (A). The claim -- that there exist agents in equilibrium that satisfy this efficiency notion -- would be evidenced with a proof (regardless of the protocol and efficiency notion) (E4.1). 

\subsubsection{Subclaim 2: Training convergence}

The second subclaim is that training has reached a local equilibrium. We can get evidence from observing that the returns from training are diminishing, and by checking that the outputs look at least roughly reasonable (E4.2).

\subsubsection{Subclaim 3: Convergence = equilibrium}

The third subclaim is that the reached local equilibrium is most likely an approximate global equilibrium of the game (E). We argue for this claim via several subclaims.  

First, while we showed above that there \textit{exist} efficient models that would reach equilibrium, we need to show that our actual training process could find them in efficient time (C4.1). Our weak evidence for this claim is that Nash equilibria for tabular Stackelberg games can be computed in polynomial (indeed linear) time, hence are tractable (although note that our game is not tabular) (E5.1). But that still leaves significant uncertainty about whether stochastic gradient descent (SGD) can reach E in realistic training time. There are reasons to think that, in the case of debate, there are bad constants or exponents (i.e.\ the time it takes to solve E is on the higher end) -- for example, translating latent knowledge acquired via heuristics into deductive arguments might require very large argument trees (D5.1). However, we are hopeful that future research can prove tighter theoretical results linking prover-estimator debate -- or other, improved debate protocols -- to realistic training methods. 

Second, after showing that it is \textit{tractable} for our training process to reach E, we need to show that it is the \textit{most} likely equilibrium. We claim that the main failure mode here is exploration hacking (C4.2). This is mostly a hunch -- there are other possible failure modes, though we would guess that they are less important -- so we leave the evidence box blank (E5.2). One failure mode is that SGD hits a local minimum and never reaches E (D5.3). We do not currently have good evidence against this defeater, though it might be possible for fields like learning theory to generate relevant evidence. Another failure mode is gradient hacking, i.e.\ using knowledge of its loss landscape to manipulate the training process (D5.2).\footnote{If the models are gradient hacking, it could in theory make it impossible to reach E via the training process T. Gradient hacking is therefore also a defeater to E5.1.} Since the system covered by this safety case is assumed to not be \textit{extremely} superhuman, we think it is reasonable to assume it will not be capable of gradient hacking. 

Third, having limited the problem to exploration hacking, we need evidence to rule out this failure mode (C4.3). As described above, this requires \textit{exploration guarantees} which provide evidence that the system has exhausted high-reward exploration options, such that convergent behaviour is likely to be a result of actually reaching E (E5.3). Note that exploration guarantees are much more useful if we are confident in our reward function (as the next key claim argues). If our reward function is bad, more exploration does not necessarily lead to better outcomes. We do not currently know how to get robust exploration guarantees - while we are optimistic about this research direction, this is a major gap in our current sketch. 

\subsection{Key claim 2: The game incentivises correctness}

The second key claim is that in the approximate global equilibrium of the game, the system is approximately honest -- its error rate is at most $\varepsilon$ (C2.2).

\begin{figure}[h]
    \centering
    \includegraphics[width=\textwidth]{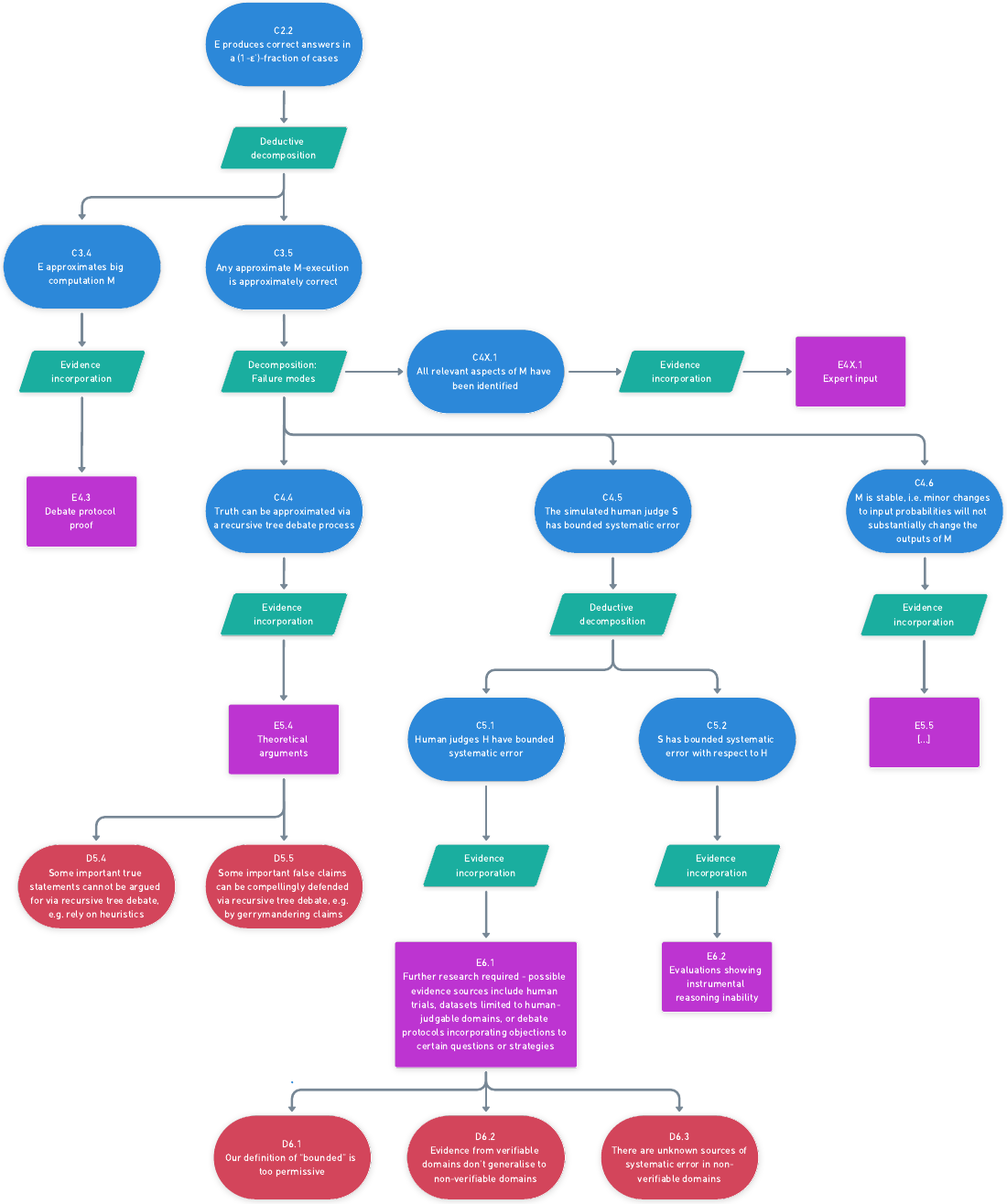}
    \caption{{The second key claim (C2.2) and the argument required to support it, in CAE notation.}}
    \label{fig:keyclaim2}
    \vspace{-10pt}
\end{figure}

Figure \ref{fig:keyclaim2} shows how this key claim is broken down.

The key claim is supported by two subclaims: 

\begin{enumerate}
    \item In E, system outputs are the same as slightly perturbed outputs of a hypothetical truth-finding process M (C3.4).
    \item For most tasks, all slightly perturbed outputs of M are true (C3.5).
\end{enumerate}

Why go this indirect route via M? Because we cannot directly verify if the system's outputs are true - that's the reason we need scalable oversight in the first place.\footnote{While we can probably directly verify many of the model's outputs in some domains (e.g.\ maths), we also want to be able to use the model to answer non-verifiable questions.} Instead of checking outputs against \textit{ground-truth labels}, we check the whole debate protocol against a \textit{process} we hypothesize to produce true outputs, called M. M is too expensive to run; the goal of our debate protocol is to accelerate it. If our protocol produces the same outputs as M, we can assume it produces true outputs. 

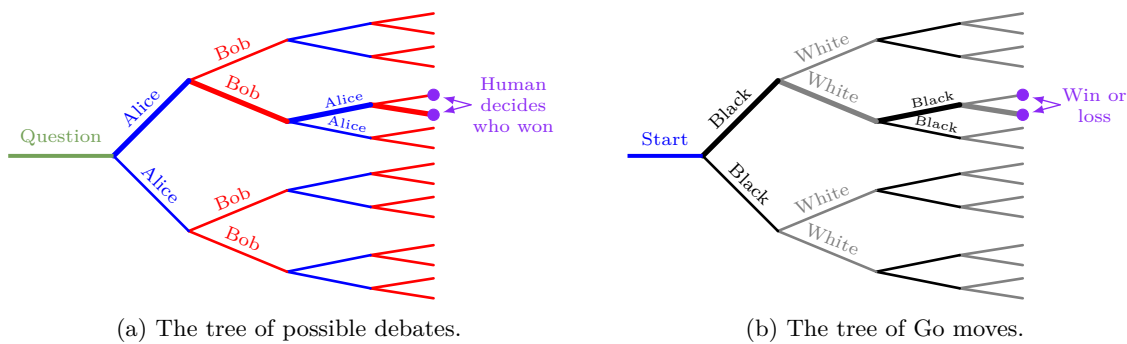
\begin{figure}[h]
\centering
\begin{minipage}[b]{.48\columnwidth}
\centering
\begin{tikzpicture}
\draw[line cap=round,red,line width=1pt] (3.4161948411340948,-1.7619156035725514) -- (4.254276189134547,-1.894654648822036) ;
\draw[line cap=round,red,line width=1pt] (3.4161948411340948,-1.7619156035725514) -- (4.254276189134547,-1.6291765583230668) ;
\draw[line cap=round,blue,line width=1pt] (2.3065629648763766,-1.541196100146197) -- (3.4161948411340948,-1.7619156035725514) ;
\draw[line cap=round,red,line width=1pt] (3.4161948411340948,-1.3204765967198424) -- (4.254276189134547,-1.453215641969327) ;
\draw[line cap=round,red,line width=1pt] (3.4161948411340948,-1.3204765967198424) -- (4.254276189134547,-1.1877375514703579) ;
\draw[line cap=round,blue,line width=1pt] (2.3065629648763766,-1.541196100146197) -- (3.4161948411340948,-1.3204765967198424) ;
\draw[line cap=round,red,line width=1pt] (1.0000000000000002,-1.0) -- (2.3065629648763766,-1.541196100146197) node[red,midway,sloped,above=-0.1em] {\scriptsize Bob};
\draw[line cap=round,red,line width=1pt] (3.4161948411340948,-0.6795234032801574) -- (4.254276189134547,-0.8122624485296419) ;
\draw[line cap=round,red,line width=1pt] (3.4161948411340948,-0.6795234032801574) -- (4.254276189134547,-0.5467843580306728) ;
\draw[line cap=round,blue,line width=1pt] (2.3065629648763766,-0.458803899853803) -- (3.4161948411340948,-0.6795234032801574) ;
\draw[line cap=round,red,line width=1pt] (3.4161948411340948,-0.23808439642744855) -- (4.254276189134547,-0.37082344167693315) ;
\draw[line cap=round,red,line width=1pt] (3.4161948411340948,-0.23808439642744855) -- (4.254276189134547,-0.10534535117796398) ;
\draw[line cap=round,blue,line width=1pt] (2.3065629648763766,-0.458803899853803) -- (3.4161948411340948,-0.23808439642744855) ;
\draw[line cap=round,red,line width=1pt] (1.0000000000000002,-1.0) -- (2.3065629648763766,-0.458803899853803) node[red,midway,sloped,above=-0.1em] {\scriptsize Bob};
\draw[line cap=round,blue,line width=1pt] (0,0) -- (1.0000000000000002,-1.0) node[blue,midway,sloped,above=-0.1em] {\scriptsize Alice};
\draw[line cap=round,red,line width=1pt] (3.4161948411340948,0.23808439642744855) -- (4.254276189134547,0.10534535117796398) ;
\draw[line cap=round,red,line width=1pt] (3.4161948411340948,0.23808439642744855) -- (4.254276189134547,0.37082344167693315) ;
\draw[line cap=round,blue,line width=1pt] (2.3065629648763766,0.458803899853803) -- (3.4161948411340948,0.23808439642744855) node[blue,pos=.7,sloped,above=-0.2em] {\tiny Alice};
\draw[line cap=round,red,line width=2pt] (3.4161948411340948,0.6795234032801574) -- (4.254276189134547,0.5467843580306728) ;
\draw[line cap=round,red,line width=1pt] (3.4161948411340948,0.6795234032801574) -- (4.254276189134547,0.8122624485296419) ;
\draw[line cap=round,blue,line width=2pt] (2.3065629648763766,0.458803899853803) -- (3.4161948411340948,0.6795234032801574) node[blue,pos=.7,sloped,above=-0.2em] {\tiny Alice};
\draw[line cap=round,red,line width=2pt] (1.0000000000000002,1.0) -- (2.3065629648763766,0.458803899853803) node[red,midway,sloped,above=-0.1em] {\scriptsize Bob};
\draw[line cap=round,red,line width=1pt] (3.4161948411340948,1.3204765967198424) -- (4.254276189134547,1.1877375514703579) ;
\draw[line cap=round,red,line width=1pt] (3.4161948411340948,1.3204765967198424) -- (4.254276189134547,1.453215641969327) ;
\draw[line cap=round,blue,line width=1pt] (2.3065629648763766,1.541196100146197) -- (3.4161948411340948,1.3204765967198424) ;
\draw[line cap=round,red,line width=1pt] (3.4161948411340948,1.7619156035725514) -- (4.254276189134547,1.6291765583230668) ;
\draw[line cap=round,red,line width=1pt] (3.4161948411340948,1.7619156035725514) -- (4.254276189134547,1.894654648822036) ;
\draw[line cap=round,blue,line width=1pt] (2.3065629648763766,1.541196100146197) -- (3.4161948411340948,1.7619156035725514) ;
\draw[line cap=round,red,line width=1pt] (1.0000000000000002,1.0) -- (2.3065629648763766,1.541196100146197) node[red,midway,sloped,above=-0.1em] {\scriptsize Bob};
\draw[line cap=round,blue,line width=2pt] (0,0) -- (1.0000000000000002,1.0) node[blue,midway,sloped,above=-0.1em] {\scriptsize Alice};
\definecolor{human}{RGB}{139,40,245};
\definecolor{question}{RGB}{116,161,88};
\node[human,right,font=\scriptsize,align=center] (end) at (4.652493324883001,0.6795234032801574) {Human\\decides\\who won};
\draw[human,fill=human] (4.254276189134547,0.5467843580306728) circle (.075) node (leaf0) {};
\draw[human,line cap=round,line width=.5pt,-latex] (4.752493324883001,0.6795234032801574) -- (leaf0);
\draw[human,fill=human] (4.254276189134547,0.8122624485296419) circle (.075) node (leaf1) {};
\draw[human,line cap=round,line width=.5pt,-latex] (4.752493324883001,0.6795234032801574) -- (leaf1);
\draw[question,font=\scriptsize,line width=1.5pt,-] (-1.4,0)--(0,0) node[above,midway] {Question};
\end{tikzpicture}
\subcaption{The tree of possible debates.}
\end{minipage}%
\begin{minipage}[b]{.48\columnwidth}
\centering
\begin{tikzpicture}
\draw[line cap=round,gray,line width=1pt] (3.4161948411340948,-1.7619156035725514) -- (4.254276189134547,-1.894654648822036) ;
\draw[line cap=round,gray,line width=1pt] (3.4161948411340948,-1.7619156035725514) -- (4.254276189134547,-1.6291765583230668) ;
\draw[line cap=round,black,line width=1pt] (2.3065629648763766,-1.541196100146197) -- (3.4161948411340948,-1.7619156035725514) ;
\draw[line cap=round,gray,line width=1pt] (3.4161948411340948,-1.3204765967198424) -- (4.254276189134547,-1.453215641969327) ;
\draw[line cap=round,gray,line width=1pt] (3.4161948411340948,-1.3204765967198424) -- (4.254276189134547,-1.1877375514703579) ;
\draw[line cap=round,black,line width=1pt] (2.3065629648763766,-1.541196100146197) -- (3.4161948411340948,-1.3204765967198424) ;
\draw[line cap=round,gray,line width=1pt] (1.0000000000000002,-1.0) -- (2.3065629648763766,-1.541196100146197) node[gray,midway,sloped,above=-0.1em] {\scriptsize White};
\draw[line cap=round,gray,line width=1pt] (3.4161948411340948,-0.6795234032801574) -- (4.254276189134547,-0.8122624485296419) ;
\draw[line cap=round,gray,line width=1pt] (3.4161948411340948,-0.6795234032801574) -- (4.254276189134547,-0.5467843580306728) ;
\draw[line cap=round,black,line width=1pt] (2.3065629648763766,-0.458803899853803) -- (3.4161948411340948,-0.6795234032801574) ;
\draw[line cap=round,gray,line width=1pt] (3.4161948411340948,-0.23808439642744855) -- (4.254276189134547,-0.37082344167693315) ;
\draw[line cap=round,gray,line width=1pt] (3.4161948411340948,-0.23808439642744855) -- (4.254276189134547,-0.10534535117796398) ;
\draw[line cap=round,black,line width=1pt] (2.3065629648763766,-0.458803899853803) -- (3.4161948411340948,-0.23808439642744855) ;
\draw[line cap=round,gray,line width=1pt] (1.0000000000000002,-1.0) -- (2.3065629648763766,-0.458803899853803) node[gray,midway,sloped,above=-0.1em] {\scriptsize White};
\draw[line cap=round,black,line width=1pt] (0,0) -- (1.0000000000000002,-1.0) node[black,midway,sloped,above=-0.1em] {\scriptsize Black};
\draw[line cap=round,gray,line width=1pt] (3.4161948411340948,0.23808439642744855) -- (4.254276189134547,0.10534535117796398) ;
\draw[line cap=round,gray,line width=1pt] (3.4161948411340948,0.23808439642744855) -- (4.254276189134547,0.37082344167693315) ;
\draw[line cap=round,black,line width=1pt] (2.3065629648763766,0.458803899853803) -- (3.4161948411340948,0.23808439642744855) node[black,pos=.7,sloped,above=-0.2em] {\tiny Black};
\draw[line cap=round,gray,line width=2pt] (3.4161948411340948,0.6795234032801574) -- (4.254276189134547,0.5467843580306728) ;
\draw[line cap=round,gray,line width=1pt] (3.4161948411340948,0.6795234032801574) -- (4.254276189134547,0.8122624485296419) ;
\draw[line cap=round,black,line width=2pt] (2.3065629648763766,0.458803899853803) -- (3.4161948411340948,0.6795234032801574) node[black,pos=.7,sloped,above=-0.2em] {\tiny Black};
\draw[line cap=round,gray,line width=2pt] (1.0000000000000002,1.0) -- (2.3065629648763766,0.458803899853803) node[gray,midway,sloped,above=-0.1em] {\scriptsize White};
\draw[line cap=round,gray,line width=1pt] (3.4161948411340948,1.3204765967198424) -- (4.254276189134547,1.1877375514703579) ;
\draw[line cap=round,gray,line width=1pt] (3.4161948411340948,1.3204765967198424) -- (4.254276189134547,1.453215641969327) ;
\draw[line cap=round,black,line width=1pt] (2.3065629648763766,1.541196100146197) -- (3.4161948411340948,1.3204765967198424) ;
\draw[line cap=round,gray,line width=1pt] (3.4161948411340948,1.7619156035725514) -- (4.254276189134547,1.6291765583230668) ;
\draw[line cap=round,gray,line width=1pt] (3.4161948411340948,1.7619156035725514) -- (4.254276189134547,1.894654648822036) ;
\draw[line cap=round,black,line width=1pt] (2.3065629648763766,1.541196100146197) -- (3.4161948411340948,1.7619156035725514) ;
\draw[line cap=round,gray,line width=1pt] (1.0000000000000002,1.0) -- (2.3065629648763766,1.541196100146197) node[gray,midway,sloped,above=-0.1em] {\scriptsize White};
\draw[line cap=round,black,line width=2pt] (0,0) -- (1.0000000000000002,1.0) node[black,midway,sloped,above=-0.1em] {\scriptsize Black};
\definecolor{human}{RGB}{139,40,245};
\node[human,right,font=\scriptsize,align=center] (end) at (4.652493324883001,0.6795234032801574) {Win or\\loss};
\draw[human,fill=human] (4.254276189134547,0.5467843580306728) circle (.075) node (leaf0) {};
\draw[human,line cap=round,line width=.5pt,-latex] (4.752493324883001,0.6795234032801574) -- (leaf0);
\draw[human,fill=human] (4.254276189134547,0.8122624485296419) circle (.075) node (leaf1) {};
\draw[human,line cap=round,line width=.5pt,-latex] (4.752493324883001,0.6795234032801574) -- (leaf1);
\draw[blue,font=\scriptsize,line width=1.5pt,-] (-1,0)--(0,0) node[above,midway] {Start};
\end{tikzpicture}
\subcaption{The tree of Go moves.}
\end{minipage}
\caption{An analogy between debate and go, represented visually: in both cases a single game with strong
players is evidence as to which moves were best. Reproduced from \cite{irving2018ai}.\label{fig:tree}}
\end{figure}

To understand M, consider a debate as a branching tree of arguments and counterarguments. A comprehensive debate would expand on every possible argument and counterargument, and have a judge consider every branch. Recursive debate aims to accelerate this process by having debaters expand only on a single path through the tree. If the debaters are good, we should expect them to anticipate how the debate will proceed given various arguments they could make and pick the optimal path through the tree. The outcome of a single debate along a single path thereby provides evidence about which debater would have won, had every argument and counterargument been considered. This is analogous to how a single game of Go provides evidence as to who the better player is, even though it only traces one path through the tree of possible moves (see Figure \ref{fig:tree}). 

Our M is equivalent to expanding the whole tree. It is a computation which produces every possible argument and counterargument at every stage and calls (simulated) human judges to adjudicate every individual path through the tree. We hypothesize that such a computation would produce correct outputs within the question domain. The goal of our debate protocol is to produce the same outputs, but more efficiently.  

\subsubsection{Subclaim 1: M-approximation}

The first subclaim is that we have succeeded at producing a protocol that accelerates M. More precisely: System outputs in E are the same as slightly perturbed output of M. In practice, our protocol cannot exactly reproduce M, since it is intractable for training to converge to a precise equilibrium - we can only reach an approximate equilibrium. As such, we can only reproduce a slightly perturbed version of M, rather than M exactly.  

Based on previous research on debate, we know that the primary failure mode for this claim is the obfuscated arguments problem \citep{barnes2020debate}. As described above, an obfuscated argument is one where both debaters know the argument contains a flaw, but it is computationally intractable for either debater to identify the flaw. If obfuscated arguments are a winning strategy, our protocol will produce different outcomes to M: expanding the whole debate would expose the obfuscated argument as false, but given computational limitations, debate might fail to get the same result. 

We are optimistic that a new debate protocol, prover-estimator debate (forthcoming), can evade the obfuscated arguments problem. Regardless of which debate protocol is used, the safety case should contain a proof that it successfully approximates M (E4.3). 

\subsubsection{Subclaim 2: Truth of M}

The second subclaim is that we have succeeded at identifying an M which produces mostly true outputs. More precisely: for most tasks, all slightly perturbed outputs of M are true. We identify three key potential issues - with a side-claim that there are no additional issues (C4X.1), supported by evidence that enough red-teaming has occurred (E4X.1).  

First, we need to argue that the \textbf{structure} of recursive tree debate is conducive to identifying truth (C4.4). This is supported by theoretical arguments (E5.4). We might worry that some true statements cannot be argued for compellingly via deductive arguments. For example, some truths might only be knowable via heuristics learned by encountering many examples, reducing the scope of questions that can be answered with recursive tree debate (D5.4). This might be an especially plausible defeater in the AI research domain, where research taste and plans may be hard to defend deductively. More worryingly, we might worry that some false claims can be compellingly defended via deductive arguments. For example, it might be possible to define the same term differently in different subclaims, rendering an argument that appears valid but is not; if human judges are called only at the leaf nodes, such inconsistencies might not be caught by M (D5.5). Some debate protocols, such as cross-examination, avoid this failure mode, but prover-estimator debate -- our example in this safety case -- does not. However, we are hopeful that future debate protocols can combine elements of both. 

Second, we need to argue that the \textbf{inputs} into M are accurate, i.e.\ that simulated human judges are competent at assessing the outcomes of debate (C4.5). Random errors should wash out over the course of training, so we are only worried about systematic error, for example due to claims playing on human biases or emotions. We need two claims here: that the human input being simulated has bounded systematic error with respect to reality (C5.1) and that the simulation has bounded systematic error with respect to the human input (C5.2). 

We imagine that the first claim could be supported with empirical results, such as trial debates (E6.1). In verifiable domains like math, human judgement could be compared to ground truth (though this may non generalise to non-verifiable domains (D6.2)). In non-verifiable domains, we could at least check if human judgement looks roughly reasonable or falls prey to known biases (though there might still be biases we are unaware of (D6.3)). Another challenge is to know what level of systematic error is ``low enough'' (D6.1). 

For the second claim about the accuracy of the simulation, the primary failure mode we have in mind is scheming to deliberately misrepresent human judgement -- perhaps in collusion with the system we are training. We imagine this claim could be supported with an inability argument based on evaluations on capabilities such as instrumental reasoning and deception (E6.2), though it is an open question if such an argument will go through for sufficiently capable judge models. However, even if not, we are optimistic that the measures used to prevent scheming in the remainder of the safety case will also apply to judge models.

We expect that getting good human inputs will be one of the most challenging parts of this safety case and require much more work. While some errors like moral or emotional bias are likely less prevalent in the domain of AI R\&D, there will likely still be errors like cognitive biases and reasoning failures. However, we are tentatively optimistic that human inputs can be improved, both by designing better processes (e.g.\ giving humans access to the right tools, finding the right number and type of judges) and by improving debate protocols (e.g.\ allowing debaters to block certain types of claims on which humans are likely to be systematically wrong). 

Third, since training does not reproduce M exactly, but rather a slightly perturbed version of M, we need to argue that this approximation is not substantially different (C4.6). This is known as stability and can be more precisely stated as: small variations in input probabilities will not change the outcome of overall computation. We are not yet sure how to support this claim, so we leave the evidence box blank (E5.5). 

We expect that most protocols which avoid obfuscated arguments -- including prover-estimator debate - will only approximate M and so require stability. However, it may not be needed for all protocols. For example, it might be possible to relax this assumption if we aim only for honesty rather than accuracy (more on this to come in future work). Alternatively, it might be possible to transform an unstable computation M into a stable computation M$*$; if so, the safety case should instead contain a claim that the outputs of M$*$ are (approximately) identical to the outputs of M).  

Even if stability is required to produce useful outputs, it might not be strictly necessary for safety. For example, in prover-estimator debate, unstable queries return no answer, rendering the protocol safe for all queries but useful only for stable queries. 

\subsection{Key claim 3: The system's behaviour will stay similar during deployment}

The third key claim is that the system's error rate will not drift past $\varepsilon$ during deployment (C2.3).

\begin{wrapfigure}{r}{0.5\textwidth}
    \includegraphics[height=0.37\textwidth]{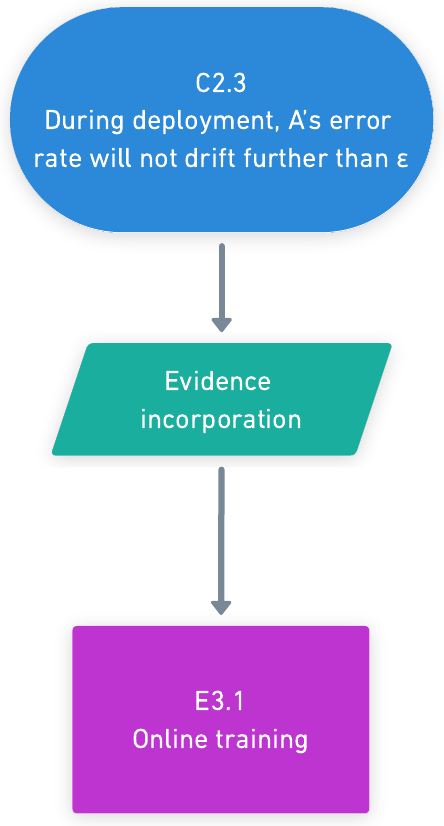}
    \centering
    \caption{{The third key claim (C2.3) and the argument required to support it, in CAE notation.}}
    \label{fig:keyclaim3}
    \vspace{-10pt}
\end{wrapfigure}

Figure \ref{fig:keyclaim3} shows how this key claim is broken down.

In our sketch, we support this claim by using online training throughout deployment as above, such that any distributional shifts will be accounted for via further training (E3.1). While the argument is straightforward, the evidence should still demonstrate that the online training regime has been correctly configured. For example, the latency (i.e.\ frequency of updates) should be high enough to prevent significant deviations from E, and there should be an effective process for updating multiple copies of the same model.  

This is one of the key claims in the sketch which breaks in a high-stakes context, where even a small number of errors can cause large harm. Since we cannot use online training, we instead need guarantees against alignment faking (and, less importantly, more generic distribution shifts). This would be much trickier to defend; see some initial ideas in ``Extending the safety case to high-stakes contexts''. 

\subsection{Key claim 4: It is sufficient for safety purposes for the system to provide correct answers most of the time}

\begin{figure}[h]
    \centering
    \includegraphics[width=0.8\textwidth]{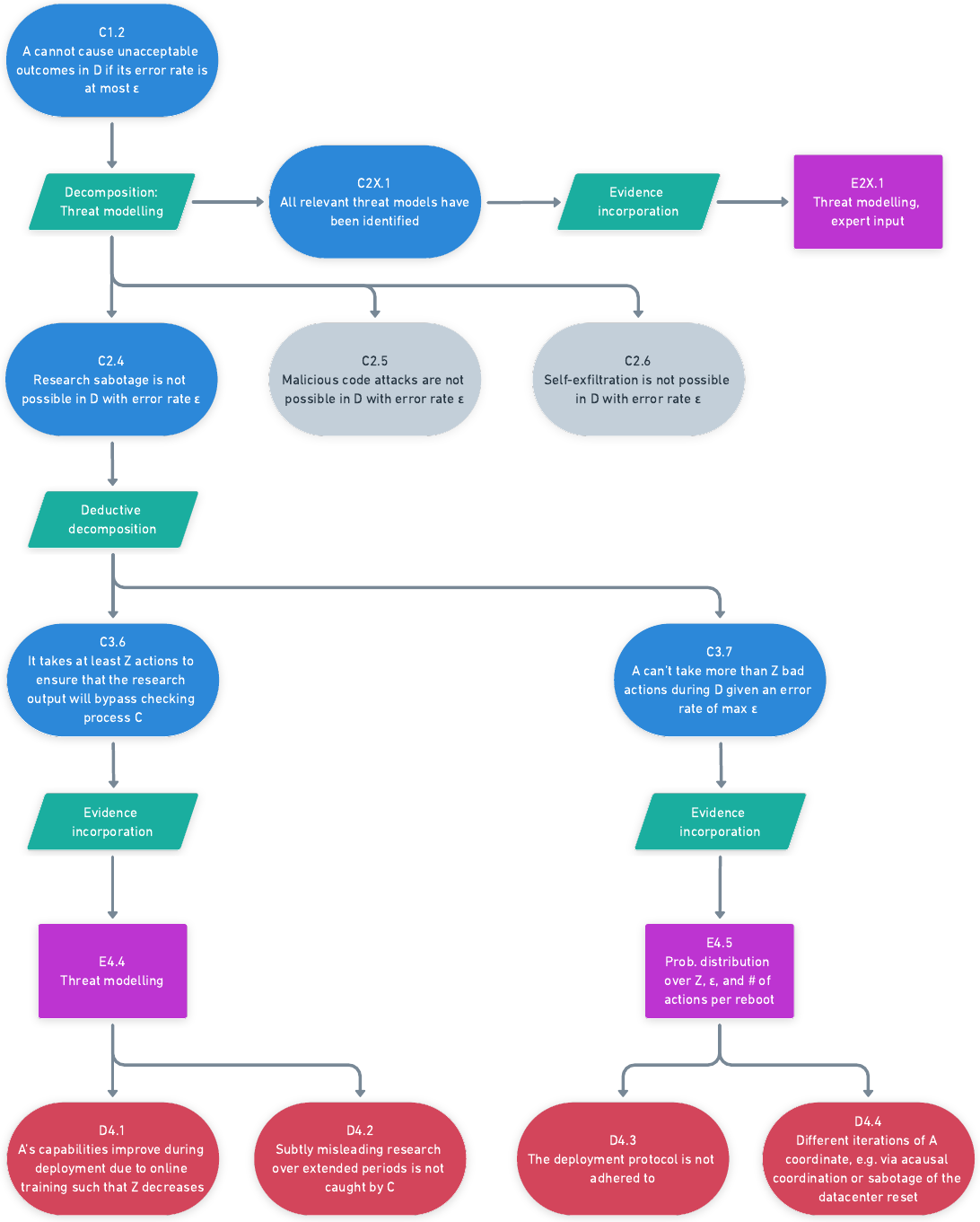}
    \caption{{The fourth key claim (C1.2) and the argument required to support it, in CAE notation.}}
    \label{fig:keyclaim4}
    \vspace{-10pt}
\end{figure}

The fourth key claim is that the system cannot cause unacceptable outcomes given an error rate of $\varepsilon$ (C1.2). Figure \ref{fig:keyclaim4} shows how this key claim is broken down. 

The main failure mode we consider is that the system is scheming to produce harmful outputs specifically in cases where our reward signal (i.e.\ human judging on debate leaf nodes) cannot catch it. The remaining errors could then constitute a coordinated set of actions that might collectively cause an unacceptable outcome.  

Since we do not have a good way to tell where our reward signal fails and what the effects of those errors would be, we conservatively assume that each error results in the worst-case outcome, equivalent to the system operating under maximal adversarial competence.\footnote{In this sense, our argument has a ``control'' flavour, in that we are assuming that the system is as egregiously misaligned as possible given the constraints of training. An alternative argument would be that we are in a low-stakes context because the reward signal catches the most severe errors.} We then argue that the system's deployment context is nevertheless low-stakes because the actions available to the system are not harmful enough to cause unacceptable outcomes in a small number of steps.\footnote{For simplicity, we phrase these claims deterministically rather than probabilistically, and with respect to a binary acceptable vs. unacceptable outcome rather than utility loss. For a continuous version, our overall claim could be that we have a regret bound of $O(\rho k\sqrt{T} + \rho \varepsilon' T)$. For a probabilistic version, we could use probability distributions over $\varepsilon$, as well as $Z$ or $\rho$.}

First, we break down the unacceptable outcomes of concern into three threat models: research sabotage, i.e.\ producing false research papers (C2.4), malicious code attacks, i.e.\ exfiltrating malware when information is extracted from the offline datacentre (C2.5), and self-exfiltration, i.e.\ the agent escaping from the datacentre (C2.6). This breakdown is supported with a side-claim that we have identified all relevant failure modes (C2X.1) - we validate this with threat modelling supported by expert input (E2X.1). The reasoning behind our breakdown is as follows: Our deployment set-up is such that the agent can only affect the external world via outputs from the datacentre. So, the only way the agent can only cause harm via those outputs, or by self-exfiltrating. Within the outputs, the agent can cause harm either via legitimate outputs (papers) or illegitimate outputs (malicious code).  

For simplicity, we only cover one threat model -- research sabotage via paper outputs - in the remainder of the safety case, though a similar argument (or other arguments) could be made for other threat models. We argue for this via two subclaims: 

\begin{enumerate}
    \item Committing effective research sabotage would require a certain number of bad actions, $Z$ (C3.6)
    \item The agent cannot take that many bad actions during a single iteration given its error rate (which is maintained via online training) (C3.7)
\end{enumerate}

\subsubsection{Subclaim 1: $Z$ bad actions required  }

The first subclaim is that committing effective research sabotage would require a certain number of bad actions, $Z$. Research sabotage could be actions such as faking experimental data, running misleading experimental set-ups, or including false claims or bugged code in the final research paper. Our deployment set-up aims to make it harder to commit research sabotage by investigating the outputted research paper with a checker model trained on n debates. To successfully conduct research sabotage, the agent would have to produce a paper that is false in a way that would very robustly not be caught by the reward signal (or very robustly manipulate the human judges).\footnote{The agent would also have to set this up using only actions that would not be flagged by the reward signal as suspicious, since it is subject to online training while producing the paper.} Our sketch argues that this is not possible, via threat modelling (E4.4).  

One potential challenge for such threat modelling is that system capabilities might improve over the course of deployment due to online training. This could enable the system to access new, more harmful actions that could enable research sabotage with fewer actions (D4.1). Another defeater is that the possibility of subtle research sabotage over long periods of time, for example by including subtly misleading but not technically false statements that are not caught by the reward signal (D4.2). This risk is reduced by resetting the datacentre between each iteration, minimising coordination opportunities. This could be supplemented with other measures to prevent suspicious correlations between outputs over long time horizons.

\subsubsection{Subclaim 2: Low likelihood of $Z$ bad actions}

The second claim is that the agent cannot take $Z$ bad actions during a single iteration given its error rate. This is straightforward to support, based on a calculation of the expected number of bad actions a system can take during deployment given probability distributions over its error rate $\varepsilon$ (given online training latency $k$), total number of actions $n$, and actions required to pull off research sabotage $Z$ (E4.5). Ideally, the claim should also be supported with evidence that the deployment protocol will in fact be adhered to, for example by committing to regular external verification (D4.3).\footnote{It may also be necessary to provide evidence that the AI agent itself will not tamper with the deployment protocol, for example by disabling online training.} The argument only works if each iteration is isolated and cannot coordinate with other iterations -- otherwise, the errors required to commit research sabotage could accumulate across iterations. While we aim to avoid this failure mode by resetting the agent and datacentre between each iteration, systems could interfere with this resetting mechanism -- or they may be able to coordinate acausally due to being based on the same underlying system (D4.4).

\section{Discussion}

\subsection{Extending the safety case to high-stakes contexts}

This safety case sketch only works in low-stakes contexts where it takes many bad actions to cause unacceptable outcomes. Key claim 4 -- that we are in a low-stakes context -- breaks by definition. Key claim 3 -- that the error rate will stay consistent throughout deployment -- also breaks because we can no longer rely on online training, opening up the possibility of alignment faking and other distribution shift problems. 

But the low-stakes claim probably will not hold for the most dangerous superhuman systems. What, then, is the value of this safety case sketch?

The most straightforward reason is some highly consequential deployments might plausibly be low-stakes. For example, early-AGI systems in sufficiently careful conditions could plausibly be low-stakes and used for essential alignment research to prevent future, high-stakes systems from causing a catastrophe. It might be possible to build a robust automated alignment safety case based on the scalable oversight techniques in this sketch -- we are planning to do more work on this. 

More indirectly, we think this safety case sketch is a useful starting point for developing other high-stakes safety case sketches. In particular, we think we might be able to extend the techniques and evidence in this sketch to high-stakes contexts, rather than requiring completely different kinds of techniques or evidence. We currently have two hopes in this direction: 

First, we might be able to use online training-like techniques during inference, on candidate actions instead of on final actions. This would allow us to maintain an error rate of $\varepsilon$ for candidate actions, then cut the regret further by generating many candidate actions. This might be enough to patch key claim 3; it is not enough to completely patch key claim 4, but it gets us some of the way. 

Second, the evidence needed to support key claim 1 and 2 might be sufficient (or extendable) to patch key claims 3 and 4. For example, to address exploration hacking, we might need a method for eliciting bad contexts (e.g.\ based on interpretability or singular learning theory). Such a method might also be useful for generating datasets that mimic deployment sufficiently effectively that we can minimise distributional shifts. More broadly, it seems like, given a theorem showing that the error rate is at most $\varepsilon$, in practice one should put significant weight on something like full intent alignment (or at least the system not being egregiously misaligned).  For example, it seems plausible that human error is anticorrelated with the severity of error such that the errors that remain are inconsequential. 

We are far from confident in these techniques, but we think it is worth trying hard to make them work. The UK AI Security Institute's research agenda encompasses further work in this area \citep{AISI2025}.

\subsection{Open problems}

Our safety case sketch faces significant limitations but also helps us identify concrete research directions that might address them. We identify five main areas for further research: 

\textbf{Improved debate protocols. }Many gaps in our safety case might be solvable with improved debate protocols with the following properties: 

\begin{itemize}
    \item \textbf{Avoids obfuscated arguments}: We are optimistic that prover-estimator debate (forthcoming) can avoid obfuscated arguments, but it remains an open question.
    \item \textbf{Computationally efficient}: The lower the complexity of reaching equilibrium in our debate game, the more likely that models that reach equilibrium are findable via stochastic gradient descent or other realistic training methods. For prover-estimator debate, it is unclear if the debaters can be constructed with polynomial-time algorithms -- further research could confirm this or develop more efficient protocols.
    \item \textbf{Does not require stability}: We are optimistic that the obfuscated arguments problem can be evaded by assuming stability. But stability might not hold for relevant queries. Future work could modify prover-estimator debate to relax the stability assumption, for example by focusing on honesty instead of correctness or by transforming an unstable computation into a stable one.
    \item \textbf{Avoids gerrymandering concerns}: Debate protocols such as cross-examination punish debaters for inconsistency across claims at the same recursive level. Others, such as, prover-estimator debate, do not. It may be possible to combine these protocols (or find other solutions).
    \item \textbf{Avoids systematic human error}: Debate protocols could be designed to minimise the impact of systematic human error, for example by allowing debaters to block queries on which they think humans are likely to be biased.
\end{itemize}

\textbf{Empirical testing and optimisation}. While debate has theoretical appeal, we have only limited data on its practical effectiveness \citep{arnesen2024training,barnes2020writeup,kenton2024scalable,khan2024debating,michael2023debate}. For example, most previous experiments use the QuALITY dataset \citep{pang2022quality} and simple debate protocols with low recursive depth. Key questions include whether debate measurably improves performance, whether training reliably reaches equilibrium, whether stability is a real problem, and whether obfuscated arguments actually emerge as predicted. We also need an improved understanding of systematic human errors -- for example: How good are humans at giving correct answers in verifiable domains, what are common systematic errors and in which domains are they most prominent, and how we can design human input for optimal results? A final set of questions surrounds the deployment set-up. For example, what counts as an "action" for the purpose of online learning in the datacentre? And will many questions in the relevant domain be difficult to answer via deductive arguments?

\textbf{Exploration hacking.} A major gap in our sketch is the requirement for exploration guarantees. We know very little about how serious a problem exploration hacking is and how to prevent it - yet the sketch depends on avoiding it. We outlined various research directions that might help address exploration hacking across both interpretability, singular learning theory, and more prosaic techniques -- much more work is needed to develop and test these. More foundationally, it would be useful to develop model organisms of exploration hacking to help test potential solutions. For very superhuman systems, we might also need to address gradient hacking. 

\textbf{High-stakes contexts.} In the high-stakes context, we face greater generalisation challenges. We lose the ability to use online training, so need to face the possibility of distribution shifts and alignment faking; and we need new ways to contend with coordinated errors not captured by our reward signal. We outline research directions in the previous section. 

\textbf{From honesty to alignment.} We need to better understand the alignment value of honest question-answering. How much does honesty plus absence of alignment faking translate into aligned actions during alignment research? How can deployment environments be designed to minimize risk while maintaining the system's usefulness for alignment research? 

The UK AI Security Institute's research agenda encompasses further work in this area \citep{AISI2025}.

\section{Conclusion}

Overall, the argument in this sketch is not strong with current techniques. Much better techniques are needed before developers can rely on alignment arguments. That said, we think there are useful directions for future research. 

We are optimistic that safety case sketches are a useful tool for research and coordination. Writing this sketch helped us clarify our thinking and gave us ideas for further research directions, forming the basis of the alignment section of the UK AI Security Institute's research agenda \citep{AISI2025}.

\begin{appendices}
    \section{The full safety case diagram}

\begin{landscape}
    \begin{figure}[h!]
        \centering
        \includegraphics[width=1.44\textwidth]{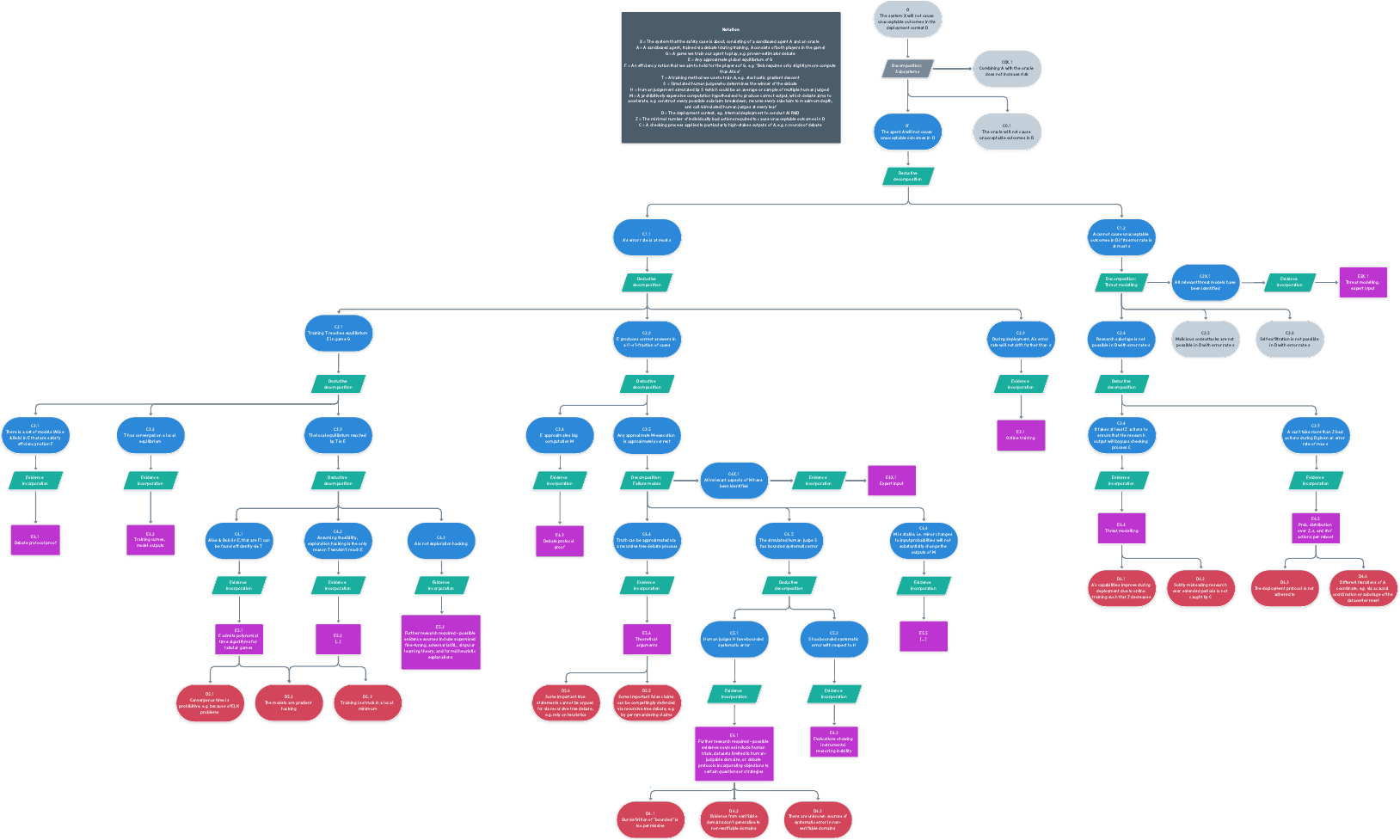}
        \caption{The full CAE diagram showing the entire safety case sketch.}
        \label{fig:completesketch}
    \end{figure}
\end{landscape}

    Figure \ref{fig:completesketch} shows the CAE diagram in its entirety.
    \section{Claims Arguments Evidence (CAE) notation}

The safety case sketch begins with a top-level claim we want to make about the safety properties of an AI system (A). This top-level claim is iteratively broken down into further subclaims, which must eventually bottom out in evidence. These subclaims shows the logical structure and dependencies connecting evidence to the top-level claim. 

The diagram uses the ``Claims, Arguments, Evidence'' notion format \citep{bloomfield2014building}. It consists of four types of units: 

\textbf{Blue ovals are claims.} Claims must either be broken down into further subclaims or directly supported by evidence, in both cases via an argument. 

\textbf{Green parallelograms are arguments.} Arguments link claims to evidence or further subclaims. They explain why the evidence or subclaims suffice to justify the claim. This sketch uses 3 argument types:  

\begin{itemize}
    \item \textbf{Decomposition}: Breaks down a claim into several subclaims. This must be supported by a ``side-claim'' explaining why the subclaims are adequate to justify the claim - unless the subclaims logically imply the claim, in which case it is called a ``deductive decomposition'' and can be free-standing. (Our sketch also uses ``deductive decompositions'' in cases where the subclaims do not strictly imply the claim but could be rephrased to do so.)
    \item \textbf{Substitution}: Replaces a claim with another single subclaim which is easier to support. This must be supported by a ``side-claim'' explaining why the subclaim adequately captures the important parts of the original claim.
    \item \textbf{Evidence incorporation}: Explains why the evidence adequately supports a given claim.
\end{itemize}

\textbf{Purple boxes are evidence. }They provide information, data or calculations that support claims. There are many possible types of evidence, including results from empirical tests, proofs, results from expert surveys, or theoretical outputs such as threat modelling reports. 

\textbf{Red ovals are defeaters.} These are hypothetical claims that, if true, could invalidate or render insufficient the evidence provided. The safety case should explain what has been done to address defeaters and/or why they are not a concern. 

\textbf{Grey elements provide context.} They are surrounding parts that would be included in a complete safety case but are not the focus of this sketch. 
\label{appdx:cae}
\end{appendices}

\section*{Acknowledgments}

We are grateful for valuable input and comments from Beth Barnes, Charlie Griffin, Jacob Hilton, Jonah Brown-Cohen, Joshua Clymer, Mart\'in Soto, Nicola Ding, and Will Kirby. 

\bibliographystyle{tmlr}
\bibliography{main}

\end{document}